\documentclass[sigconf,nonacm]{acmart}
\AtBeginDocument{%
  }

\setcopyright{acmlicensed}
\copyrightyear{2018}
\acmYear{2018}
\acmDOI{XXXXXXX.XXXXXXX}
\acmConference[Conference acronym 'XX]{Make sure to enter the correct
  conference title from your rights confirmation email}{June 03--05,
  2018}{Woodstock, NY}
\acmISBN{978-1-4503-XXXX-X/2018/06}

\begin{document}

\title{World-Coordinate Human Motion Retargeting via SAM 3D Body}

\author{Zhangzheng Tu}
\authornote{Both authors contributed equally to this research.} 
\affiliation{%
  \institution{Dalian University of Technology}
  \country{China}
}
\email{tuzhangzheng@mail.dlut.edu.cn}

\author{Kailun Su}
\authornotemark[1] 
\affiliation{%
  \institution{Shenzhen University}
  \country{China}
}
\email{kaslensu@gmail.com}

\author{Shaolong Zhu}
\affiliation{%
  \institution{Harbin Institute of Technology, Shenzhen}
  \country{China}
}
\email{zhushaolong2004@gmail.com}

\author{Yukun Zheng}
\affiliation{%
  \institution{ShenZhen University}
  \country{China}
}
\email{zhengyukun2005@gmail.com}


\begin{abstract}
Recovering \emph{world-coordinate} human motion from monocular videos with humanoid robot retargeting is significant for embodied intelligence and robotics.
To avoid complex SLAM pipelines~\cite{shen2024world, shin2024wham, wang2024tram} or heavy temporal models~\cite{goel2023humans}, we propose a lightweight, engineering-oriented framework that \textbf{leverages SAM 3D Body (3DB)}~\cite{yang2025sam3d} as a frozen perception backbone and uses the \textbf{Momentum Human Rig (MHR)}~\cite{ferguson2025mhr} representation as a robot-friendly intermediate.

Our method (i) locks the identity and skeleton-scale parameters of per tracked subject to enforce temporally consistent bone lengths, (ii) smooths per-frame predictions via efficient sliding-window optimization in the low-dimensional MHR~\cite{ferguson2025mhr} latent space, and (iii) recovers physically plausible global root trajectories with a differentiable soft foot--ground contact model and contact-aware global optimization.
Finally, we retarget the reconstructed motion to the Unitree G1 humanoid using a kinematics-aware two-stage inverse kinematics pipeline.

Results on real monocular videos show that our method has stable world trajectories and reliable robot retargeting, indicating that structured human representations with lightweight physical constraints can yield robot-ready motion from monocular input.

\end{abstract}



\begin{teaserfigure}
    \centering
  \includegraphics[width=0.85\textwidth]{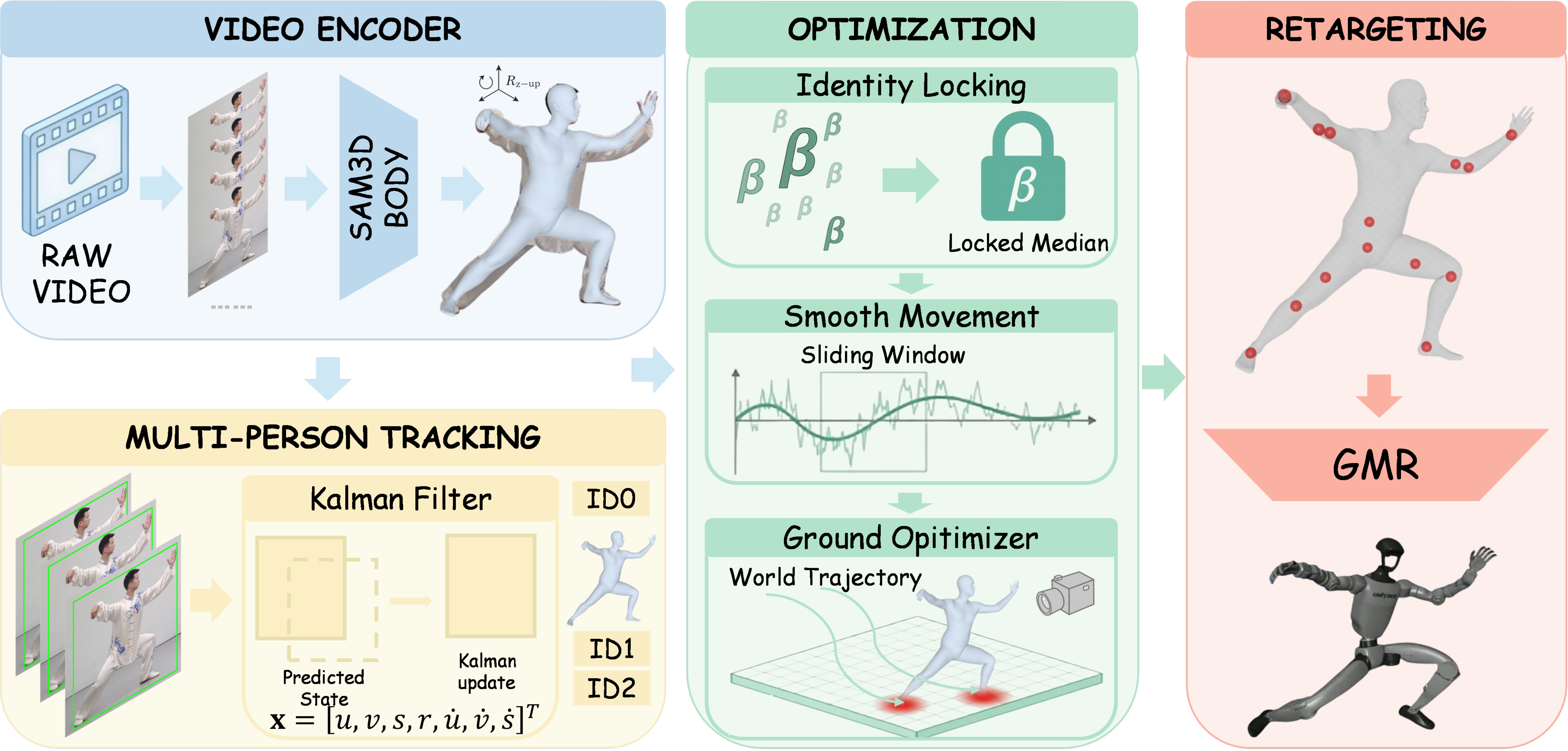}
  \caption{
Overview of the proposed pipeline for world-coordinate human motion retargeting from monocular video.
Raw video frames are processed by SAM 3D Body (3DB)~\cite{yang2025sam3d}, used as a frozen backbone, to extract per-frame MHR parameters~\cite{ferguson2025mhr}. A detect–track module with Kalman filtering associates identities over time. Trajectory-level identity and scale locking are then applied, followed by sliding-window latent-space smoothing to reduce high-frequency jitter. A contact-aware ground optimizer estimates physically plausible root trajectories in the world coordinate system. Finally, the reconstructed motion is retargeted to the Unitree G1 humanoid robot via a kinematics-aware retargeting pipeline.
}
  \label{fig:teaser}
  \Description{}
\end{teaserfigure}


\maketitle

\section{Introduction}

Understanding and transferring human motion from visual observations to humanoid robots is a long-standing goal in embodied intelligence.
Monocular videos are particularly attractive as an input modality due to their low cost and wide availability.
However, recovering \emph{world-coordinate} and physically consistent human motion from monocular input still remains challenging~\cite{shen2024world,shin2024wham,wang2024tram,li2022d}, especially when driving real humanoid robots rather than achieving visually accurate reconstruction.

Recent advances in monocular 3D human reconstruction have significantly improved per-frame pose and shape estimation.
Yet, most methods operate primarily in the camera coordinate system~\cite{pavlakos2018learning,fiche2025mega,kanazawa2018end,zhang2023pymaf} and focus on visual fidelity, leaving global translation, metric scale, and contact dynamics either under-constrained or implicitly handled.
Approaches that recover world trajectories often rely on complex SLAM systems or heavy temporal models~\cite{kocabas2020vibe}, introducing substantial engineering overhead and limiting practical deployment in robot retargeting scenarios.

In this work, we explore a complementary, engineering-oriented direction, as illustrated in Fig.~\ref{fig:teaser}.
Instead of proposing a new human body model or learning a large temporal network, we investigate how a \emph{structured human representation} combined with lightweight physical constraints can yield robot-ready motion from monocular videos.
Our framework leverages \textbf{SAM 3D body (3DB)}~\cite{yang2025sam3d} as a frozen perception backbone and uses the \textbf{Momentum Human Rig (MHR)}~\cite{ferguson2025mhr} representation as an explicit kinematic intermediate between vision and robot control.

The key observation underlying our approach is that MHR~\cite{ferguson2025mhr} provides temporally consistent skeletal structure with invariant bone lengths, which aligns naturally with the rigid-link assumptions of humanoid robot kinematics.
Building on this property, we design a pipeline that enforces trajectory-level identity and scale consistency, suppresses high-frequency pose jitter via latent-space optimization, and recovers physically plausible world-coordinate root trajectories using contact-aware global optimization.
Importantly, these steps avoid the need for full SLAM~\cite{loper2023smpl} or complex scene reconstruction, while remaining differentiable and computationally lightweight.

We demonstrate the effectiveness of our approach by retargeting~\cite{araujo2025retargeting} reconstructed motions to the Unitree G1 humanoid robot.
Experiments on real-world monocular videos show that our method produces stable world trajectories and enables reliable motion execution on the robot.
Our results suggest that combining off-the-shelf human reconstruction with structured kinematics and minimal physical constraints offers a practical pathway toward robot-ready motion understanding from monocular vision.

In summary, this work makes the following contributions:
(i) an engineering-oriented pipeline that leverages 3DB for world-coordinate human motion retargeting,
(ii) a trajectory-level identity and skeleton-scale locking strategy for kinematic consistency,
(iii) a contact-aware global optimization scheme for physically plausible root motion, and
(iv) a complete retargeting system demonstrated on a real humanoid robot.

\section{Related Work}

\subsection{Evolution of Parametric Human Models: From SMPL to MHR}

Parametric human body models are a central tool in 3D vision, among which the \textbf{Skinned Multi-Person Linear (SMPL)} model~\cite{loper2023smpl} has become the de facto standard for pose estimation, motion capture, and animation. SMPL parameterizes human shape and pose and produces articulated meshes via Linear Blend Skinning (LBS)~\cite{kavan2007skinning}, offering an efficient and differentiable representation widely adopted by monocular and multi-view reconstruction methods.

Despite its success, SMPL is less suited for robotics-oriented motion reasoning. Because joint locations depend on body shape, bone lengths vary across identities and poses, violating rigid-link kinematic assumptions. Moreover, the tight coupling between skeletal structure and surface deformation complicates enforcing temporal kinematic consistency, limiting its effectiveness as an intermediate representation for humanoid motion retargeting.

To address these issues, MHR~\cite{ferguson2025mhr} explicitly decouples skeletal kinematics from surface geometry. By preserving temporally invariant bone lengths for a fixed identity while retaining expressive surface deformation, MHR provides stable joint hierarchies and differentiable kinematic solvers, making it well suited for world-coordinate motion recovery and robot retargeting.

\subsection{World-Coordinate Reconstruction from Monocular Human Motion}

Recovering metrically scaled, world-coordinate human motion from monocular video has been widely studied. Existing methods typically rely on temporal integration with motion or contact priors~\cite{shin2024wham}, scene-level scale recovery via SLAM or geometric cues~\cite{wang2024tram}, or canonical coordinate systems that mitigate long-term integration drift~\cite{shen2024world}. While these approaches improve global consistency, they often introduce substantial engineering complexity, depend on reliable static scene cues, or require heavy temporal modeling.

In contrast, our work adopts a lightweight, engineering-oriented approach. Rather than pursuing high-precision scene reconstruction, we leverage a kinematics-aware intermediate representation (MHR) together with noisy camera priors and contact-aware optimization to recover world-coordinate trajectories that are sufficient for downstream humanoid retargeting. This design prioritizes robustness and practical deployability over absolute visual accuracy.

\begin{figure*}[t]
  \includegraphics[width=0.85\textwidth]{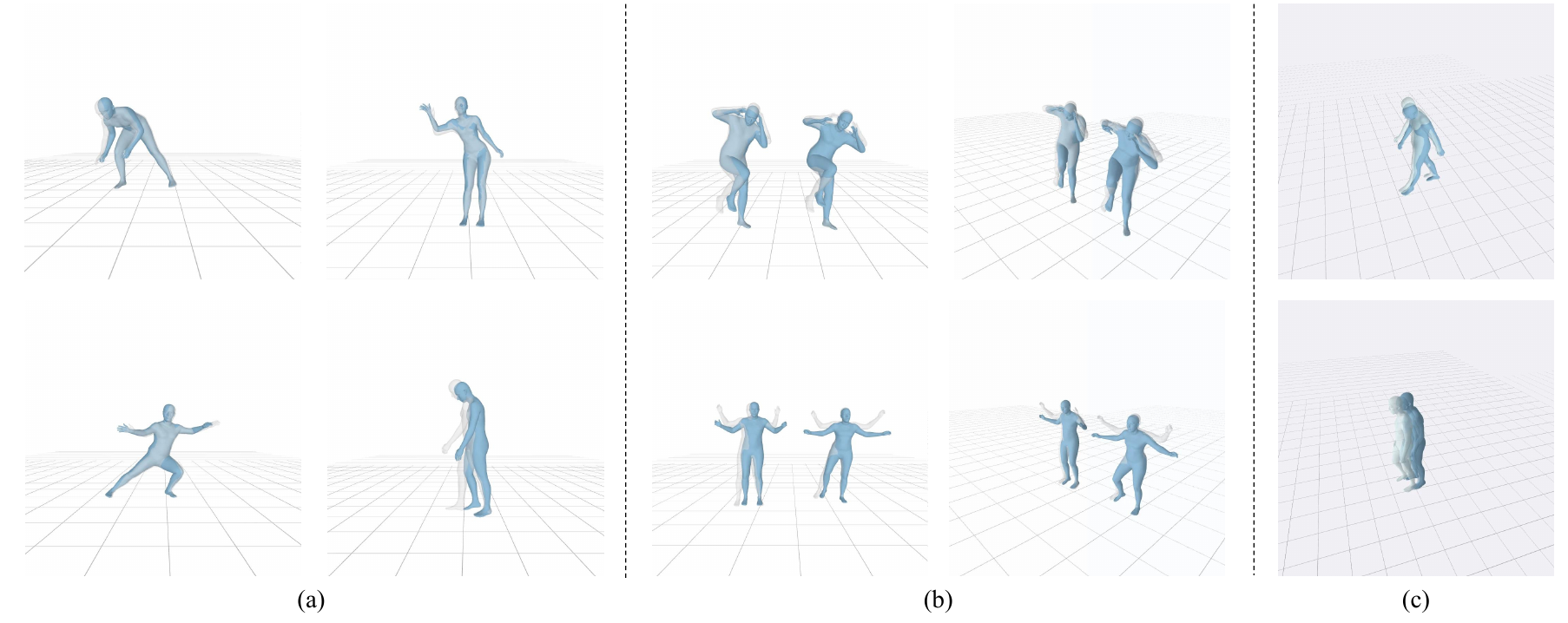}
  \caption{
Qualitative results.
(a) Temporal consistency of single-person motion estimation.
(b) Robust motion estimation under multi-person settings.
(c) Ablation on temporal and geometric refinement.
Dark blue denotes the current-frame mesh; gray and light blue denote the refined and unrefined predictions two frames ahead, respectively.
}
\Description{}
  \label{fig:qualitative_results}
\end{figure*}

\section{Method}

\subsection{Preliminaries}

\paragraph{Momentum Human Rig (MHR)}
MHR~\cite{ferguson2025mhr} is a structured and differentiable parametric human representation that explicitly disentangles \textit{identity}, \textit{expression}, and \textit{motion} factors. The model parameters represent articulated pose through a compact latent space with LBS and pose-dependent correctives. This disentanglement enables temporally consistent bone lengths, which is particularly important for kinematics-aware retargeting.

\paragraph{SAM 3D Body (3DB)}
3DB~\cite{yang2025sam3d} is a single-image full-body reconstruction system built on MHR. It employs a shared image encoder with separate decoding branches for body and hands and is trained using large-scale data generated through multi-stage 2D/3D annotation, fitting, and quality control. In our pipeline, 3DB is used as an off-the-shelf backbone without modification or retraining.

\subsection{Video Preprocessing and Tracking}
We extract frames from the input video and run 3DB~\cite{yang2025sam3d} per frame.
To maintain identity consistency under multi-person scenes, we adopt a detect--track pipeline using SORT with Kalman filtering at the 2D bounding-box level.
The tracker state is
$\mathbf{x} = [u, v, s, r, \dot{u}, \dot{v}, \dot{s}]^\top$,
where $(u,v)$ is the box center, $s$ is area, $r$ is aspect ratio, and the remaining terms are temporal derivatives.
We associate detections to tracks using the Hungarian algorithm with IoU-based costs.

\subsection{Trajectory-Level Consistency and Latent-Space Smoothing}

Frame-wise monocular reconstructions often suffer from temporal fluctuations in body shape, skeletal scale, and pose, which degrade kinematic stability and downstream retargeting quality.
Since identity-related factors should remain invariant within a short temporal track, we first enforce trajectory-level consistency on shape and scale parameters, and then perform pose smoothing directly in the low-dimensional MHR~\cite{ferguson2025mhr} latent space.

\paragraph{Trajectory-level shape and scale locking.}
Given a tracked sequence of length $T$, we aggregate per-frame predictions and compute trajectory-level descriptors by temporal averaging:
\begin{equation}
\beta_{\text{shape}}^{\text{final}}=\mathrm{mean}(\{\beta_{shape,t}\}_{t=1}^{T}),\quad
\gamma_{\text{scale}}^{\text{final}}=\mathrm{mean}(\{\gamma_{scale,t}\}_{t=1}^{T}).
\end{equation}
All frames within the same track share $\beta_{\text{shape}}^{\text{final}}$ and $\gamma_{\text{scale}}^{\text{final}}$, enforcing temporally invariant bone lengths and reducing identity drift across frames. This constraint provides a stable kinematic foundation for subsequent pose smoothing and motion retargeting.

\paragraph{Latent-space pose smoothing.}
Even after enforcing shape and scale consistency, per-frame pose estimates may contain high-frequency jitter.
We therefore perform sliding-window optimization directly in the compact MHR latent space, which allows efficient temporal refinement while preserving semantic structure.

\paragraph{Latent fidelity.}
To prevent excessive deviation from the original 3DB predictions, we constrain the optimized latents to remain close to the initial estimates:
\begin{equation}
\mathcal{L}_{\text{latent}}=
\frac{1}{T}\sum_{t}
\left(
\|z_{t}^{\text{model}}-\hat{z}_{t}^{\text{model}}\|_2^2+
\|z_{t}^{\text{expr}}-\hat{z}_{t}^{\text{expr}}\|_2^2
\right).
\end{equation}

\paragraph{Temporal smoothness.}
Let $\mathbf{p}_t^{(j)}$ and $R_t^{(j)}$ denote the global position and rotation of joint $j$ at time $t$.
We define linear and angular velocities and accelerations as:
\begin{align}
\mathbf{v}_t^{(j)} &= \mathbf{p}_{t+1}^{(j)} - \mathbf{p}_{t}^{(j)}, \quad
\mathbf{a}_t^{(j)} = \mathbf{p}_{t+2}^{(j)} - 2\mathbf{p}_{t+1}^{(j)} + \mathbf{p}_{t}^{(j)}, \\
\boldsymbol{\omega}_t^{(j)} &= \operatorname{Log}(R_{t+1}^{(j)}R_{t}^{(j)\top}), \quad
\boldsymbol{\alpha}_t^{(j)} = \boldsymbol{\omega}_{t+1}^{(j)} - \boldsymbol{\omega}_{t}^{(j)}.
\end{align}
We penalize these quantities using a generalized Charbonnier function $\rho(x;\beta,\varepsilon)=\sqrt{\beta x^2+\varepsilon}$, yielding
\begin{align}
\mathcal{L}_{\text{smooth}}=\sum_j w_j\Big(
&\lambda_v \sum_t \rho(\|\mathbf{v}_t^{(j)}\|)+
\lambda_a \sum_t \rho(\|\mathbf{a}_t^{(j)}\|)\nonumber\\
&+\lambda_\omega \sum_t \rho(\|\boldsymbol{\omega}_t^{(j)}\|)+
\lambda_\alpha \sum_t \rho(\|\boldsymbol{\alpha}_t^{(j)}\|)
\Big),
\end{align}
where joint-wise weights $w_j$ emphasize trunk and root stability while avoiding over-smoothing on highly articulated joints such as hands and face.

\paragraph{Window stitching.}
To ensure temporal continuity across sliding windows, we include a boundary consistency term $\mathcal{L}_{\text{bound}}$ that aligns overlapping frames.
The final optimization objective is:
\begin{equation}
\mathcal{L}_{\text{total}}=
\lambda_{\text{latent}}\mathcal{L}_{\text{latent}}+
\mathcal{L}_{\text{smooth}}+
\lambda_{\text{bound}}\mathcal{L}_{\text{bound}}.
\end{equation}

\subsection{Contact-Aware Global Root Optimization}
Monocular depth ambiguity makes global root translation noisy, often causing non-physical shaking and foot sliding.
We estimate a physically plausible world-coordinate root trajectory using a lightweight optimization (\textit{GroundOptimizer}) in a Z-up frame.

\paragraph{Soft contact probability.}
For each foot $f\in\{L,R\}$ with height-to-ground distance $d_f$, we compute
\begin{equation}
\begin{array}{c}
w_{\text{base}}^{f}
= \exp\!\left(-\frac{d_f^2}{2\sigma_h^2}\right), \\
\alpha^{f}
= \mathrm{softmax}\!\left(k_{\text{contact}}\, w_{\text{base}}^{f}\right),
\quad
p_c^{f} = w_{\text{base}}^{f}\, \alpha^{f}.
\end{array}
\end{equation}

yielding differentiable contact probabilities that smoothly transition between single- and double-support phases.

\paragraph{Energy formulation.}
We solve for the root translation $\mathbf{T}_{\text{global}}(t)$ by minimizing:
\begin{equation}
E_{\text{ground}}=
\lambda_{\text{phy}}\big(\mathcal{L}_{\text{slide}}+\mathcal{L}_{\text{pen}}+\mathcal{L}_{\text{contact}}\big)
+\mathcal{L}_{\text{smooth}}+\lambda_{\text{aux}}\mathcal{L}_{\text{aux}}.
\end{equation}
$\mathcal{L}_{\text{slide}}$ penalizes horizontal foot velocity during contact (reducing sliding), $\mathcal{L}_{\text{pen}}$ prevents ground penetration, and $\mathcal{L}_{\text{contact}}$ encourages feet to stay near the ground height under high $p_c$.
We also regularize root velocity and acceleration.
To reduce monocular drift, an auxiliary \emph{soft camera prior} is applied mainly in non-contact phases ($1-p_c$) with anisotropic weights to emphasize forward consistency.
We fix the first frame at the world origin and optimize using Adam.

\subsection{Retargeting to Unitree G1}
We retarget MHR motion to the Unitree G1 humanoid using a kinematics-aware pipeline.
We select 14 corresponding joints and align local coordinate conventions via:
\begin{equation}
R_{\text{final}} = R_{\text{z-up}} \cdot F \cdot R_{\text{MHR}} \cdot R_{\text{offset}}.
\end{equation}
We then apply height-ratio scaling to map motions into the robot workspace and solve IK in two stages:
(1) anchor the root and end-effectors for coarse alignment, and
(2) refine intermediate joints (e.g., knees and elbows) for feasible articulation.
This yields stable retargeted motions across heterogeneous skeletons.

\section{Qualitative Results}
Fig.~\ref{fig:qualitative_results} presents representative qualitative results of our pipeline. As shown in Fig.~\ref{fig:qualitative_results}(a), the method yields temporally consistent reconstructions across diverse single-person motions, including sports, mocap sequences, Tai Chi, and daily walking. Sliding-window latent smoothing with trajectory-level shape and scale locking suppresses frame-to-frame jitter while preserving motion-specific dynamics across both fast and slow regimes. Fig.~\ref{fig:qualitative_results}(b) shows multi-person results. The detect–track module with per-track identity locking maintains consistent bone lengths and smooth per-subject reconstructions, even under close interactions, mitigating identity switching and skeletal inconsistency. Fig.~\ref{fig:qualitative_results}(c) highlights contact-aware root optimization. Soft foot-contact constraints reduce drift and foot-sliding artifacts, producing more stable and physically plausible trajectories suitable for robot retargeting.

Overall, these results demonstrate that latent-space smoothing, trajectory-level constraints, and contact-aware optimization jointly enable temporally coherent and physically plausible 3D motion reconstruction from monocular video.

\section{Conclusion and Limitations}

We propose a lightweight, engineering-oriented pipeline that converts monocular video into world-coordinate, robot-ready human motion. Using 3DB as a frozen backbone and MHR as a compact, kinematics-aware representation, the pipeline enforces per-track skeletal consistency, reduces high-frequency jitter via sliding-window latent optimization, and recovers physically plausible root trajectories through contact-aware optimization.

The pipeline has several limitations. Monocular depth ambiguity remains challenging in multi-person scenes, often leading to incorrect relative positioning and limiting accurate modeling of interactions under occlusion. In addition, while MHR is expressive and compact, the lack of a widely accepted 4D evaluation protocol restricts comprehensive quantitative assessment.

Future work includes (1) developing MHR-specific 4D metrics to evaluate geometric accuracy, temporal smoothness, and motion plausibility, and (2) extending the pipeline for precise multi-person interactions via improved depth reasoning and multi-subject geometric constraints.

\bibliographystyle{ACM-Reference-Format}
\bibliography{software}
\end{document}